\documentclass{article}

\usepackage{microtype}
\usepackage{graphicx}
\usepackage{subfigure}
\usepackage{booktabs} 

\usepackage{hyperref}

\usepackage[accepted]{icml2019}

\usepackage{amsmath,amsfonts,bm}

\def\eqref#1{equation~\ref{#1}}

\def\1{\bm{1}}

\DeclareMathAlphabet{\mathsfit}{\encodingdefault}{\sfdefault}{m}{sl}
\SetMathAlphabet{\mathsfit}{bold}{\encodingdefault}{\sfdefault}{bx}{n}

\usepackage{url}
\usepackage{graphicx}

\usepackage{tabularx}
\usepackage{multirow}
\usepackage{verbatim}
\usepackage[title]{appendix}

\usepackage{wrapfig,booktabs}
\usepackage{xcolor}

\newcommand{\CHANGE}[1]{{#1}}

\icmltitlerunning{Learn to Grow for Overcoming Catastrophic Forgetting}

\begin{document}

\twocolumn[
\icmltitle{Learn to Grow: A Continual Structure Learning Framework for Overcoming Catastrophic Forgetting}



\icmlsetsymbol{equal}{*}
\icmlsetsymbol{intern}{$\dagger$}

\begin{icmlauthorlist}
\icmlauthor{Xilai Li}{equal,intern,to}
\icmlauthor{Yingbo Zhou}{equal,sf}
\icmlauthor{Tianfu Wu}{to}
\icmlauthor{Richard Socher}{sf}
\icmlauthor{Caiming Xiong}{sf}
\end{icmlauthorlist}

\icmlaffiliation{to}{Department of Electrical and Computer Engineering and the Visual Narrative Initiative, North Carolina State University, NC, USA.}
\icmlaffiliation{sf}{Salesforce Research, Palo Alto, CA, USA.}

\icmlcorrespondingauthor{C. Xiong}{cxiong@salesforce.com}
\icmlcorrespondingauthor{T. Wu}{tianfu\_wu@ncsu.edu}

\icmlkeywords{Learn-to-Grow, Continual Learning, Lifelong Learning, Catastrophic Forgetting, Deep Neural Networks, Machine Learning, ICML}

\vskip 0.3in
]



\printAffiliationsAndNotice{\icmlEqualContribution} 

\setlength{\textfloatsep}{6pt}

\begin{abstract}
Addressing catastrophic forgetting is one of the key challenges in continual learning where machine learning systems are trained with sequential or streaming tasks. 
Despite recent remarkable progress in state-of-the-art deep learning, deep neural networks (DNNs) are still plagued with the catastrophic forgetting problem. This paper presents a conceptually simple yet general and effective framework for handling catastrophic forgetting in continual learning with DNNs. The proposed method consists of two components: a neural structure optimization component and a parameter learning and/or fine-tuning component. 
By separating the explicit neural structure learning and the parameter estimation, not only is the proposed method capable of evolving neural structures in an intuitively meaningful way, but also shows strong capabilities of alleviating catastrophic forgetting in experiments. Furthermore, the proposed method outperforms all other baselines on the permuted MNIST dataset, the split CIFAR100 dataset and the Visual Domain Decathlon dataset in continual learning setting.    
\end{abstract}
\section{Introduction}
Learning different tasks continuously is a common and practical scenario that happens all through the course of human learning. The learning of new skills from new tasks usually does not have negative impact on the previously learned tasks. Furthermore, with learning multiple tasks that are highly related, it often helps to advance all related skills. However, this is commonly not the case in current machine learning with deep neural networks (DNNs). When presented a sequence of learning tasks, DNNs experiences so called ``catastrophic forgetting" problem~\citep{mccloskey1989catastrophic, ratcliff1990connectionist}, where they usually largely ``forget" previously learned tasks after trained for a new task. This is an interesting phenomenon that has attracted lots of research efforts recently.

To overcome catastrophic forgetting, approaches such as Elastic Weight Consolidation~\citep[EWC][]{kirkpatrick2017overcoming} and synaptic intelligence~\citep{zenke2017continual} introduce constraints to control parameter changes when learning new tasks. However, forgetting is still non-negligible with these approaches, especially when the number of tasks increases. Forgetting is also addressed with memory-based methods, where certain information regarding learned tasks are stored to help retaining the performance of the learned tasks~\citep[see][for example]{lopez2017gradient,sener2018active}. Additionally, there are  methods~\citep{mallya2018piggyback, rebuffi2017learning, rebuffi2018efficient, mancini2018adding} that  learn multiple domains and  completely avoid forgetting by adding a small amount of parameters while the previously estimated parameters are kept fixed. However, these models rely on a strong base network and knowledge transferability is limited mainly between two consecutive tasks. 

Most of the current approaches in continual learning with DNNs couple network structure and parameter estimation and usually apply the same model structure for all tasks. 
In this paper, \textit{we propose to explore a more intuitive and sensible approach}, that is to learn task specific model structures \emph{explicitly} while retaining model primitives sharing, decoupling from model parameter estimation\footnote{The structure that referred here is more fine-grained, such as number of layers, type of operations at each layer, etc. It does not refer to generic structure names like convolutional neural networks or recurrent neural networks.}. Different  tasks may require different structures, especially if they are not relevant, so it may not make much sense to employ the same structure in learning. For example, consider the tasks of learning digit and face recognition DNNs, the lower level layers (features) required for the two tasks are likely to be drastically different, thus entailing different overall structures that have task specific low level layers. Forcing the same structure for these tasks is likely to 
cause catastrophic forgetting for one task (e.g., digit recognition) after the other task (e.g., face recognition) is trained. 
On the other hand, if different tasks learn to explore different structures and grow their specific components accordingly, it still has the potential to share common feature layers while maximizing the performance for new tasks. 

In this paper, we present a \textbf{learn-to-grow framework} that explicitly separates the learning of model structures and the estimation of model parameters. In particular, we employ architecture search  to find the optimal structure for each of the sequential tasks. The search accounts for various options, such as sharing previous layers' parameters, introducing new parameters, and so on. After the structure is searched, the model parameters are then estimated. We found that \textit{ 1) explicitly continual structure learning makes more effective use of parameters among tasks, which leads to better performance and sensible structures for different tasks; 2) separating the structure and parameter learning significantly reduced catastrophic forgetting as compared to other baseline methods with similar model complexities. }

\section{The Proposed Learn-to-Grow Framework}
\subsection{Problem Definition of Continual Learning}
Consider a sequence of $N$ tasks, denoted by $\textbf{T}=(T_1, T_2, ..., T_N)$. Each task $T_t$ has a training dataset,  $\mathcal{D}^{(t)}_{train}=\{(x_i^{(t)}, y_i^{(t)}); i=1,\cdots, n_t\}$, where $y^{(t)}_i$ is the ground-truth annotation (e.g., a class label) and $n_{t}$ is the number of training examples. Let $\mathcal{D}_{train}=\cup_{t=1}^N\mathcal{D}_{train}^{(t)}$ be the entire training dataset for all tasks. Similarly, denote by $\mathcal{D}_{test}^{(t)}$ the test dataset for a task $T_t$. 
Denote by $f(\cdot; \Theta_t)$ the model (e.g., a DNN) in learning where $\Theta_t$ collects all learned parameters up to the current task $T_t$ (inclusive). The model gets to observe tasks from $1$ to $N$ sequentially. After the model is trained on a task $T_t$ using its training dataset $\mathcal{D}^{(t)}_{train}$, both $\mathcal{D}^{(t)}_{train}$ and $\mathcal{D}_{test}^{(t)}$ will not be available in training tasks from $T_{t+1}$ to $T_{N}$. \textit{The main objective of continual learning} is to maximize the performance of $f(\cdot;\Theta_t)$ at the task $T_t$ while minimizing the forgetting for tasks from $T_1$ to $T_{t-1}$, all being evaluated in their test datasets $\mathcal{D}_{test}^{(t')}$ ($1\leq t'\leq t$). 
Ideally, we would like to minimize the following objective function in this continual learning setting,
\begin{align}
    \mathcal{L}(\Theta_N; \mathcal{D}_{train}) &= \sum_{t=1}^{N} \mathcal{L}_t (\Theta_t; \mathcal{D}_{train}^{(t)}) \label{eq:ideal_loss}\\
     \mathcal{L}_t (\Theta_t; \mathcal{D}_{train}^{(t)})  &= \frac{1}{n_{t}}\sum_{i=1}^{n_{t}} \ell_t(f(x_i^{(t)};\Theta_t), y_i^{(t)}) \label{eq:task_loss}
\end{align}
where $\ell_t$ is the loss function for task $T_t$ (e.g., the cross-entropy loss) and the model regularizer term is omitted for notion simplicity. However, since we do not have access to all datasets at the same time, the above objective function (Eqn.~\ref{eq:ideal_loss}) can not be directly computed and then minimized. The challenge is to maintain $\sum_{t'=1}^{t-1} \mathcal{L}_{t'} (\Theta_{t'}; \mathcal{D}_{train}^{(t')})$ not to change too much without explicitly measuring it due to the streaming setting, while estimating $\Theta_t$ via minimizing Eqn.~\ref{eq:task_loss} in isolation. 

\begin{figure}[tbp!]
\centering
\includegraphics[width = 0.45\textwidth,trim={0 0.1in 0 0},clip]{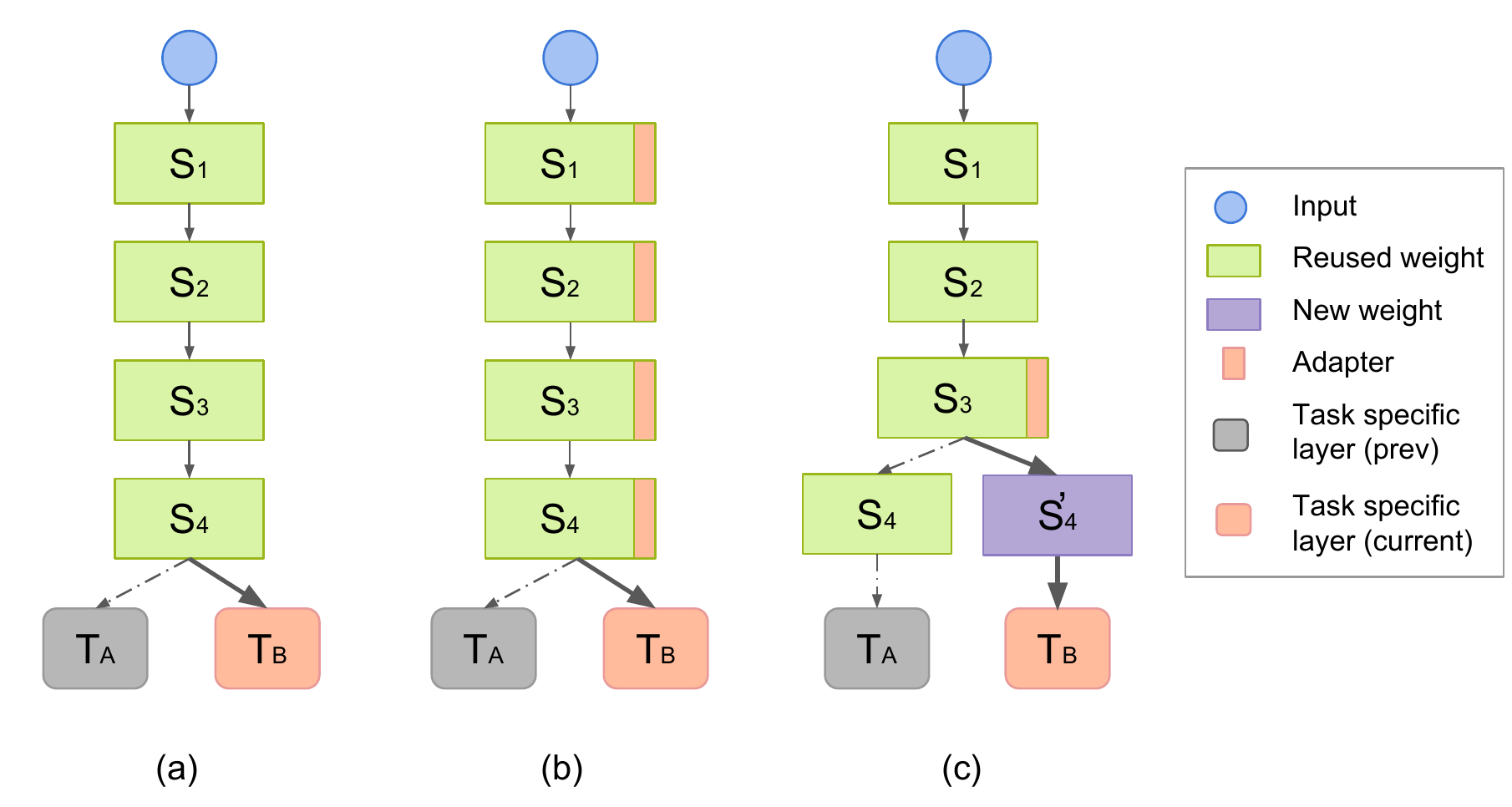}
\vspace{-10pt}
\caption{Illustration of different continual learning approaches. a) All but the task specific layer are shared, catastrophic forgetting is countered by techniques that prevents parameters to move to lower energy regions of previous tasks. b) Each task will add some fixed task specific parameters, all layers' original weights are not tuned, and thus prevents forgetting. c) Our approach, where network structure is learned by architecture search. In this example, the search result decides to ``reuse" the first two layer, do ``adaptation" for the 3rd layer and allocate
``new" weight for the 4th layer. } \label{fig:illustration}
\end{figure}

As illustrated in Fig.~\ref{fig:illustration} (b), one straightforward solution is to keep $\Theta_{t-1}$ fixed when learning $\Theta_t=\Theta_{t-1}\cup \theta_t$ to avoid catastrophic forgetting completely, where $\theta_t$ is the new parameters introduced for a new task $T_t$. How to introduce $\theta_t$ for each task sequentially is usually hand-crafted and some simple heuristics are often used, e.g., by adding some extra channels for each layer in a DNN. By doing this, the model will become more and more complicated for incoming tasks and $\Theta_{t-1}$ is ``artificially" enforced to be reused for a new task $T_t$ without accounting for their potential dissimilarities. So, the computational efficiency and accuracy performance of new tasks are traded-off for avoid catastrophic forgetting.  

As illustrated in Fig.~\ref{fig:illustration} (a), another way of addressing catastrophic forgetting is to utilize a single set of parameters $\Theta$ for all tasks, and control the changes of parameter values from $\Theta_{t-1}$ to $\Theta_t$ using some statistically inspired functions such as the Fisher information criterion used in EWC~\citep{kirkpatrick2017overcoming}. Following this direction, the accuracy performance of new tasks are usually suffered from the constrained parameter space and well-designed initial models are entailed for ensuring reasonably good performance across tasks. Furthermore, the effectiveness of the parameter change control functions is often unknown as the number of tasks increases at scale.   

\subsection{Our Proposed Learn-to-Grow Framework}
In our learn-to-grow framework (Fig.~\ref{fig:illustration} (c)), we adopt the growing strategy as stated above in learning $\Theta_t=\Theta_{t-1}\cup \theta_t$. But, we learn  to ``grow" $\theta_t$ on top the previously trained model $\Theta_{t-1}$ and to exploit $\Theta_{t-1}$ in a flexible way without enforcing to reuse all of them. Our proposed method is also complementary to the elastic parameter  strategy as used in EWC. For the learned-to-reuse parameters in $\Theta_{t-1}$, we can either keep them fixed or allow them to change subject to some elastic penalty functions. So, the proposed method can harness the best of both, and is capable of avoid catastrophic forgetting of old tasks completely without sacrificing the computational efficiency and accuracy performance of new tasks.    
We introduce $s_t(\Theta_t)$ to indicate the task-specific model for task $T_t$. The loss function (Eqn.~\ref{eq:task_loss}) is  changed to,
\begin{equation}
    \mathcal{L}_t(s_t(\Theta_t)) = \frac{1}{n_{t}}\sum_{i=1}^{n_{t}} \ell_t(f(x_i^{(t)};s_t(\Theta_t)), y_i^{(t)}) \label{eq:struct_loss}
\end{equation}
Now the structure is explicitly taken into consideration when learning all the tasks. When optimizing the updated loss function in Eqn.~\ref{eq:struct_loss}, one needs to determine the optimal parameter based on the structure $s_t$. This loss can be viewed in two ways. One can interpret it as selecting a task specific network from a `super network' that has parameter $\Theta$ using $s_t$, or for each task we train a new model with parameter $s_t(\Theta_t)$. There is a subtle difference between this two views. The former one has an constraint on the total model size, while the latter one does not. So, in the worst case scenario of the latter, the model size will grow linearly as we increase the number of tasks. This would lead to a trivial solution -- training completely different models for different tasks and is no longer continual learning! To address this problem, we propose the following penalized loss function,
\begin{align}
    \nonumber \mathcal{L}_t(s_t(\Theta_t)) = &\frac{1}{n_{t}}\sum_{i=1}^{n_{t}} \ell_t(f(x_i^{(t)};s_t(\Theta_t)), y_i^{(t)})+ \\
    &\beta_t \mathcal{R}^s_t(s_t) + \lambda_t \mathcal{R}^p_t(\Theta_t)\label{eq:struct_loss_final}
\end{align}
where $\beta_t > 0$, $\lambda_t \ge 0$, $\mathcal{R}^s_t$ and $\mathcal{R}^p_t$ represent the regularizers for the network structure and model parameters respectively. \CHANGE{For instance, one can use $\ell_2$ regularization for $\mathcal{R}^p_t$ when optimizing model parameters, and $\mathcal{R}^s_t$ can be as simple as the ($\log$) number of parameters}. In this way, the total number of parameters are  bounded from above, and the degenerate cases are thus avoided.


\section{Our Implementation}
\begin{figure*}[t]
\centering
\includegraphics[width = 1.0\textwidth,trim={0 0.1in 0 0},clip]{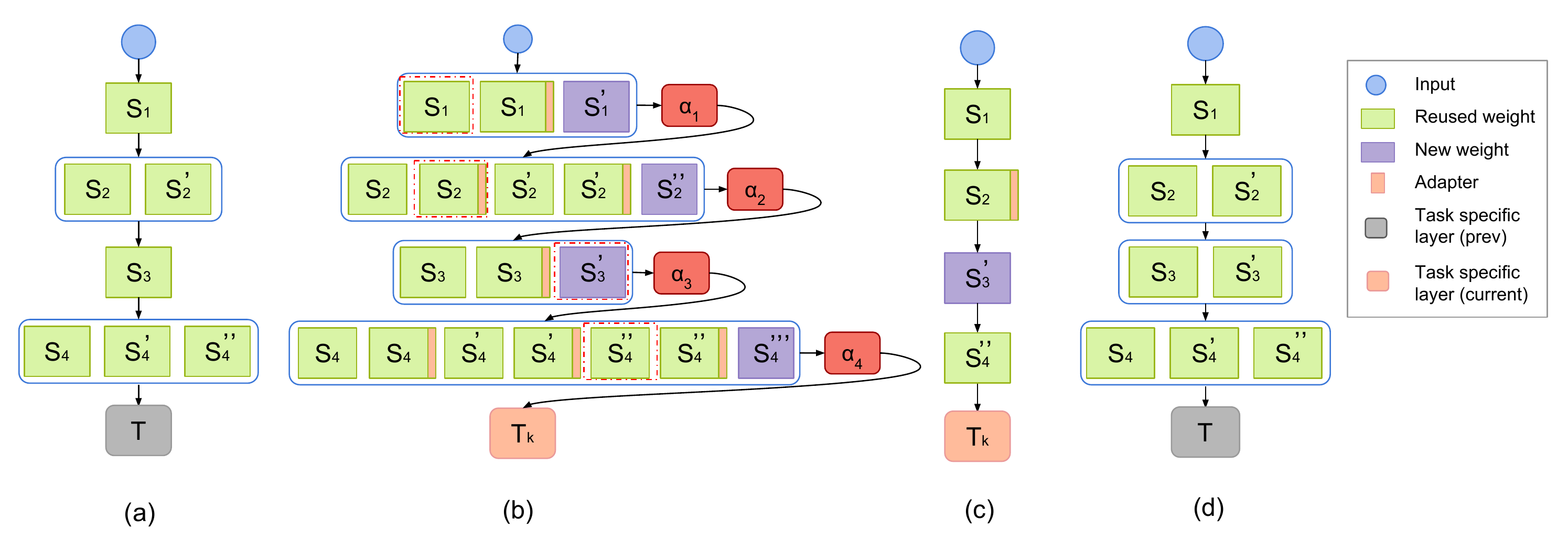}
\vspace{-20pt}
\caption{Illustration of the proposed learn-to-grow framework. a) Current state of super model. In this example, the 1st and 3rd layers have single copy of weight, while the 2nd and 4th has two and three respectively. b) During search, three options, ``{\tt reuse}", ``{\tt adaptation}"  and ``{\tt new}" are utilized. $\alpha$ is the weight parameters for the architecture. c) Parameter optimization with selected architecture on the current task k. d) Update super model to add the newly created $S'_3$. See text for details. }\label{fig:learn2grow}
\vspace{-10pt}
\end{figure*}

It is a challenging problem to optimize the loss described in Eqn.~\ref{eq:struct_loss_final}, since it involves explicit optimization of the structure of the model. 
In our implementation, we focus on continual learning with DNNs. The proposed method consists of two components: a neural structure optimization component and a parameter learning and/or fine-tuning component. The former learns the best neural structure for the current task on top of the current DNN trained with previous tasks. It learns whether to reuse or adapt building blocks or layers in the current DNN, or to create new ones if needed under the differentiable neural architecture search framework~\citep{liu2018darts}. The latter estimates parameters for newly introduced structures, and fine-tunes the old ones if preferred.
We present details in the following sections (see Fig.~\ref{fig:learn2grow}).

\subsection{Structure Optimization}
We employ neural architecture search (NAS) for structure optimization. Before we move on to further details, we adopt a further simplification that a global network topology is given and could work for all tasks of interest, such as a ResNet~\citep{he2016deep}. We optimize the wiring pattern between layers and their corresponding operations. It is straightforward to extend this to more complicated cases, e.g., by utilizing NAS at the first task. 


Consider a network with $L$ shareable layers and one task-specific layer (i.e. last layer) for each task. A super network $\mathcal{S}$ is maintained so that all the new task-specific layers and new shareable layers will be stored into $\mathcal{S}$. 

The goal of search is to seek the optimal choice for each of the $L$ layers, 
given the current task data $\mathcal{D}_{train}^{(t)}$ and all the shareable layers' weights stored in $\mathcal{S}$. The candidate
choices for each layer are defined by: ``{\tt reuse}", ``{\tt adaptation}" and ``{\tt new}". The {\tt reuse} choice will make new task use the same parameter as the previous task. The {\tt adaptation} option adds a small parameter overhead that trains an additive function to the original layer output. The {\tt new} operator will spawn new parameters of exactly the same size of the current layer parameters. Here, we denote the size of the $l_{th}$ layer in
super network $\mathcal{S}$ as $|\mathcal{S}^l|$. The total number of choices in the $l_{th}$ layer $C_l$ is $2|\mathcal{S}^l|+1$, because
we will have $|\mathcal{S}^l|$ "{\tt reuse}", $|\mathcal{S}^l|$ "{\tt adaptation}" and $l$ "{\tt new}". Thus, the total search space is
$\prod_l^L{C_l}$. One potential issue here is that, in the worst case, the search space may grow exponentially with respect to the number of tasks. One way of addressing this is to limit the total number of possible choices, and maintain a priority queue for learning the options. We do not find this necessary in all of our experiments.

Similar to DARTS~\citep{liu2018darts}, to make the search space continuous, we relax the categorical choices of the 
$l_{th}$ layer as a {\tt Softmax} over all possible $C_l$ choices, i.e.
$
x_{l+1}=\sum_{c=1}^{C_l}{\frac{\exp(\alpha_c^l)}{\sum_{c'=1}^{C_l}{\exp(\alpha_{c'}^{l})}}g_c^l(x_{l})}
$
Here, the vector $\alpha^l$ of dimension $C_l$ is the architecture weights that are used for mixing the
choices for each sharable layer. And $g_c^l$ here is the operator for the choice $c$ at layer $l$ which
is expressed as:
\begin{equation}
g_c^l(x_l)=
\left\{
	\begin{array}{ll}
		S_c^l(x_l)  & \mbox{if } c \leq |\mathcal{S}^l|, \\
		S_c^l(x_l)+\gamma_{c-|\mathcal{S}^l|}^l(x_l)  & \mbox{if } |\mathcal{S}^l| < c \leq 2|\mathcal{S}^l|, \\
		o^l(x_l) & \mbox{if } c=2|\mathcal{S}^l|+1
	\end{array}
\right.
\end{equation}
Here, $\gamma$ is the {\tt adaption} operation and $o$ the  {\tt new} operation to be trained from scratch. 
After this relaxation, the task of discrete search is posed as optimizing a 
set of continuous weights $\alpha=\left\{\alpha^l\right\}$. After the search, the optimal 
architecture is obtained by taking the index with the largest weight $\alpha_c^l$ for each layer $l$, i.e. 
$c_l=\mathop{\arg\max}\alpha^l$. 

Adopting the training strategy from DARTS, we split the training dataset $\mathcal{D}^{(t)}_{train}$ into two subsets: a validation subset for NAS, and a training subset for parameter estimation. 
We use validation loss $L_{val}$ to update the architecture  weights $\alpha$, while the parameters are estimated by the training loss $L_{train}$. 
The architecture weights and parameters are updated alternately during the search process. 
Because it is a nested bi-level optimization problem, the original DARTS provide a second-order 
approximation for more accurate optimization. In this work, we find it is sufficient to use the simple alternately 
updating approach, which was referred as the first-order approximation in \citep{liu2018darts}.

To make it clear how ``{\tt reuse}", ``{\tt adaptation}" and ``{\tt new}" operations work, we walk through a concrete example in the following. Let us take
a convolutional neural network (CNN) with all the layers using $3\times3$ kernel size as an example. The choice of 
``{\tt reuse}" is just using the existing weights and keep them fixed during learning, thus there is no additional 
parameter cost. For ``{\tt adaptation}", it uses a $1\times1$ convolution layer added to the original $3\times3$ convolution layer in parallel, similar to the adapter used in~\citep{rebuffi2017learning}.
During training, the weight of the original $3\times3$ kernel is kept fixed, while the parameters of the $1\times1$ adapter is learned.
In this case, the additional parameter cost is only $1/9$ of the original parameter size. For the ``{\tt new}" operation, it introduces a replicated $3\times3$ layer that is initialized randomly and trained from scratch. We make use of the loss function $L_{val}$ to implement the regularizer $\mathcal{R}^s_i(s_i)$. 
The value of the regularizer is set proportional to the product of the additional parameter size $z_c^l$ and its 
corresponding weight $\alpha_c^l$ (i.e. $\mathcal{R}^s_i(s_i)=\sum_{c,l}\alpha_c^l z_c^l$). 
The architecture weights $\alpha$ is optimized in terms of both
accuracy and parameter efficiency at the same time. 

\subsection{Parameter Optimization}
\label{sec:param_opt}
After the search, we retrain the model on the current task. There are two strategies to deal with ``{\tt reuse}", we can either fix it unchanged during
retraining just as in search, or fine-tune it with some regularization -- simple $\ell_2$ regularization or more sophisticated methods such as the EWC~\citep{kirkpatrick2017overcoming}. The former strategy could
avoid forgetting completely, however it will lose the chance of getting positive backward transfer, which means
the learning of new tasks may help previous tasks' performance. When the search process select ``{\tt reuse}" at layer $l$, 
it means that the $l_{th}$ layer tends to learn very similar representation as it learned from one of the previous tasks.
This indicates semantic similarity learned at this layer $l$ between the two tasks. Thus, we conjecture
that fine-tuning the ``{\tt reuse}" $l_{th}$ layer with some regularization could also benefit the previous tasks (elaborated in experiments)
After retraining on the current task, we need to update/add the created and tuned layers, 
task-specific adapters and classifiers in the maintained super network. 

\begin{figure}[!tbph]
\centering
\includegraphics[width = 0.5\textwidth,trim={0 0.1in 0 0},clip]{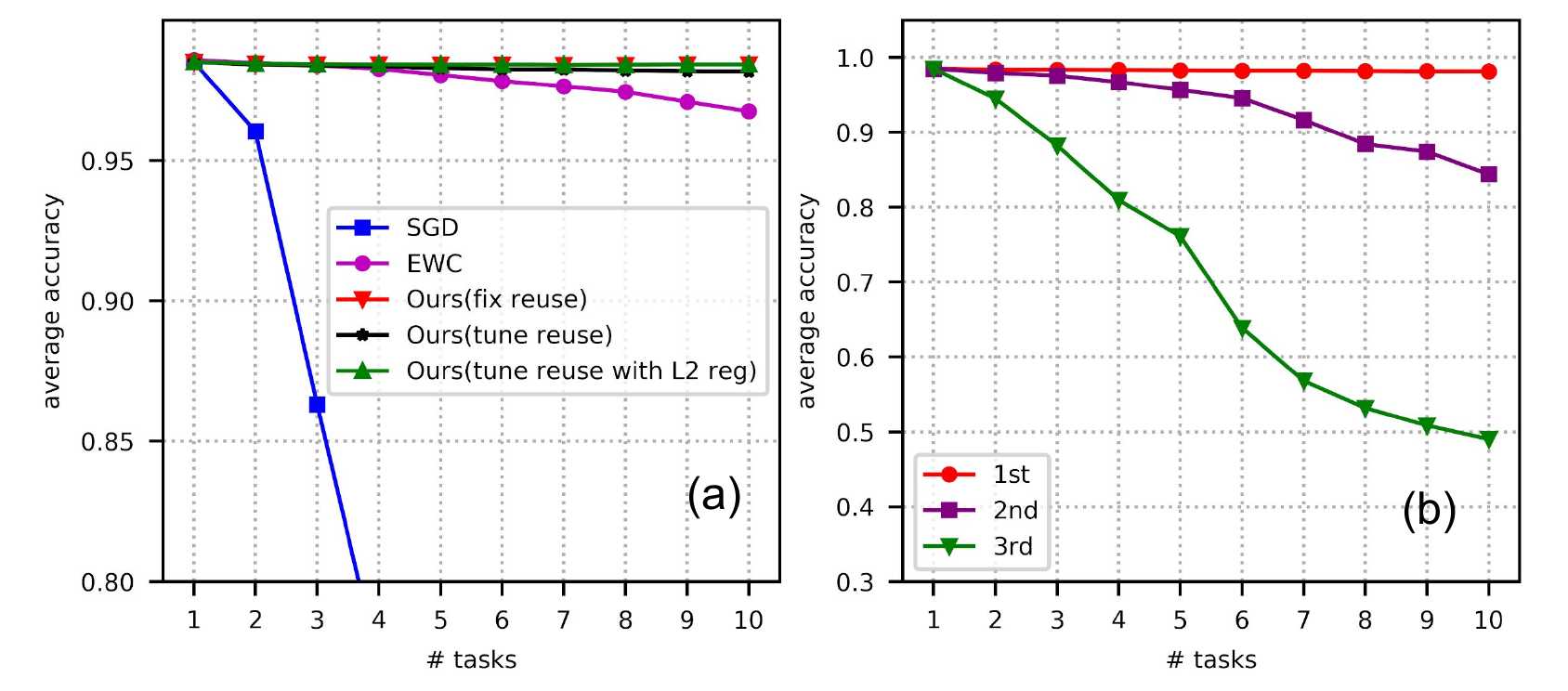}
\vspace{-0.2in}
\caption{Results on permutated MNIST dataset. 
a) Comparing our method (fix, tune reuse with and without regularization) with SGD and EWC on the average accuracy over the seen tasks. b) Ablation experiments of "new" different layers in terms of average accuracy over the seen tasks. }
\label{fig:mnist_result}
\end{figure}

\section{Experiments}
In this section, we first test two main hypotheses that motivate our proposed learn-to-grow framework and then compare with state-of-the-art continual learning approaches. First, will sensible model architectures be sought via the explicit structure learning for new tasks? Second, will the optimized structure results in better continual learning, i.e., overcoming catastrophic forgetting? We test these two hypotheses on two datasets: the permuted MNIST and the visual domain decathlon dataset~\citep{rebuffi2017learning}. \textbf{The permuted MNIST dataset} is a simple image classification problem that derived from the MNIST handwritten digit dataset~\citep{mnist}, which is commonly used as benchmark in the  continual learning literature~\citep{kirkpatrick2017overcoming, lopez2017gradient, zenke2017continual}. For each task, a unique fixed random permutation is used to shuffle the pixels of each image, while the annotated label is kept fixed. \textbf{The visual decathlon dataset} (VDD) consists of 10 image classification tasks -- ImageNet, CIFAR100, Aircraft, DPed, Textures, GTSRB, Omniglot, SVHN, UCF101 and VGG-Flowers. The images of all the tasks are rescaled with the lower-edge being $72$ pixels. The tasks are across multiple domains with highly imbalanced dataset, which makes it a good candidate to investigate the continual learning problem and analyze potential inter-task transfer, either forward or backward. 

For experiments in permuted MNIST, \CHANGE{we use a 4-layer fully-connected  networks with 3 feed-forward layers and the $4th$ layer is the \emph{shared} softmax classification layer across all tasks}. \CHANGE{This corresponds to the so called `single head' setup \citep{farquhar2018towards}. We choose to use this setting because for the permuted MNIST dataset all the tasks share the same semantics, and sharing the task classifier is a more reasonable design choice.} We test our method in learning the 10 permuted MNIST tasks in sequence. For simplicity, we only use two options, ``{\tt reuse}" and ``{\tt new}" during the structure optimization. 

For experiments in VDD, we use a 26-layer ResNet~\citep{he2016deep} as the base network to learn the first task. This network consists of 3 \textit{basic} residual blocks with output feature channels being 64, 128 and 256 respectively. 
At the end of each residual block, the feature resolution is halved by average pooling. We adopt all the three options during the search. For ``{\tt adaptation}", a $1\times1$ convolution layer is used as the adapter.


\subsection{Are Sensible Structures Sought in Learn-to-Grow?}
For the permuted MNIST dataset, we may expect that a sensible evolving architecture for the $10$ tasks tends to share higher level layers due to the same task semantics, but to differ at lower layers accounting for the pixel shuffling. 
Interestingly, our experiments show that the structure optimization indeed selects the same wiring pattern that applies ``{\tt new} to the first layer and ``{\tt reuse} to the remaining layers in all the 10 tasks. 
This shows that the structure optimization component is working properly. 

Although the learned wiring patterns are intuitive, we perform further experiments to address what if we force to use ``{\tt new}  for the learned ``{\tt reuse"} layers? We enumerate and compare the three settings that the $i$-th layer is ``{\tt new}" and others are ``{\tt reuse}" ($i=1,2,3$). In the results, we found that the learned pattern is actually the best choice compared with the other two settings (see Fig.~\ref{fig:mnist_result} b). 

\begin{figure}[t]
\centering
\includegraphics[width = 0.35\textwidth,trim={0 0.1in 0 0},clip]{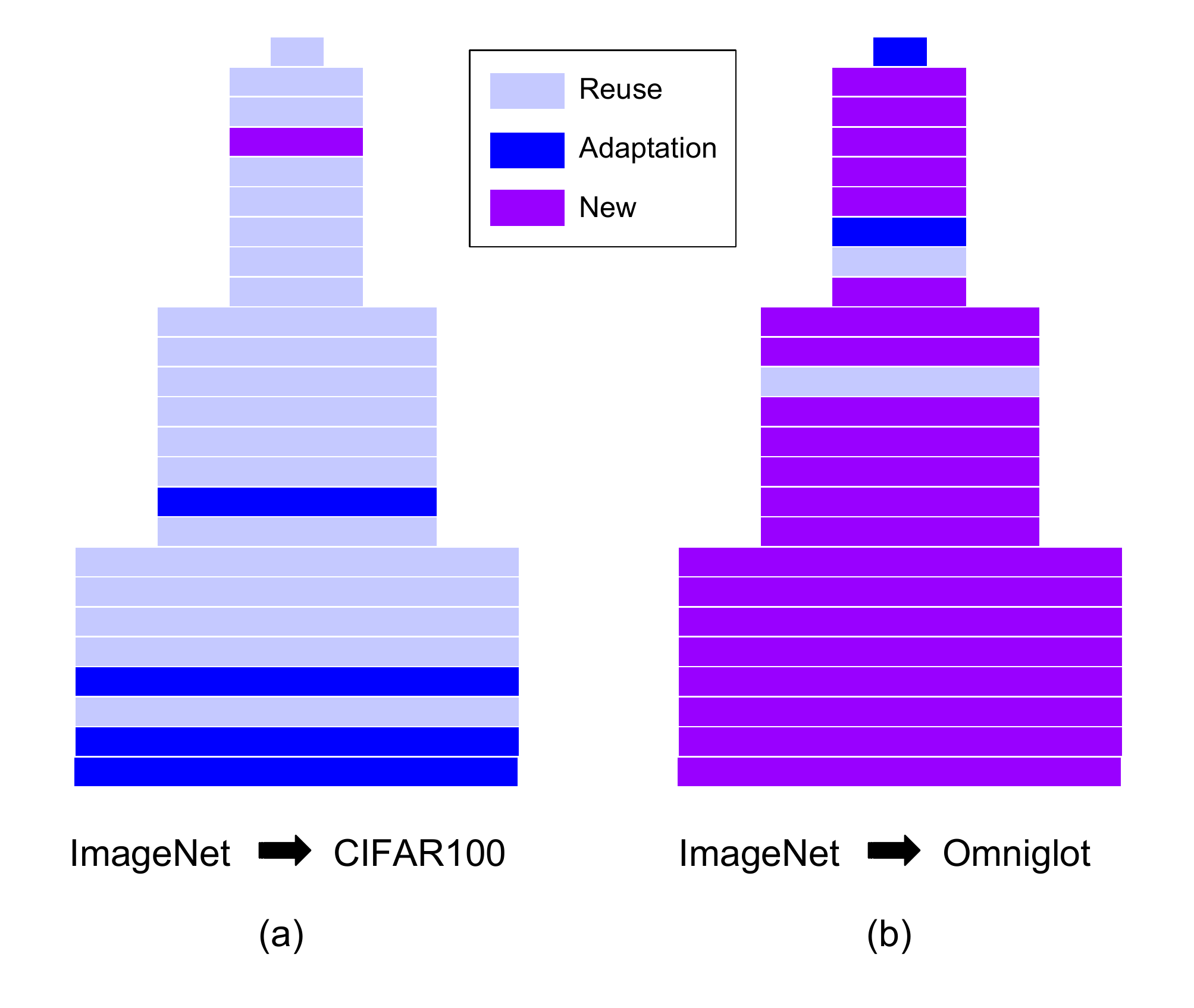}
\vspace{-10pt}
\caption{Visualization of searched architecture with learning two tasks sequentially. The search are based on the super model obtained by training ImageNet as first task. (a) and (b) shows searched architecture on CIFAR100 and Omniglot task respectively. }
\label{fig:vdd_struct}
\end{figure}

In VDD, we test our method between two tasks.  
As shown in Fig.~\ref{fig:vdd_struct} (a), when the two tasks are similar (ImageNet and CIFAR-100, both consisting of natural images), most of the layers are shared for these two tasks. When two drastically different tasks are used, e.g., ImageNet and Omniglot, as Fig.~\ref{fig:vdd_struct} (b) shows, most layers in the learned structure select the ``{\tt new}" option. 

The above experimental results show that the proposed structure optimization does result in sensible task specific structures with proper model primitive sharing. The learned task structure tends to share when the semantic representation of corresponding layers are similar, and spawn new parameters when the required information is very different.

\subsection{Are Forgetting Addressed in Learn-to-Grow?}

Obviously, if the ``{\tt reuse}'' layers are kept fixed in learning, our method does not have any forgetting. We are interest in how significant the forgetting will be when we fine-tune the ``{\tt reuse"} layers.

We first test this in the permuted MNIST. As a baseline, we show that simply updating all the layers with stochastic gradient descent (SGD) from task to task (i.e., the setting in Fig.~\ref{fig:illustration} (a)) results in catastrophic forgetting (see Fig.~\ref{fig:mnist_result} (a)). After training the 10 tasks sequentially, the average accuracy dropped from 97.9\% to 63.0\%. With the EWC used in learning~\citep{kirkpatrick2017overcoming}, the forgetting is alleviated and the average accuracy is 96.8\%. 
For our proposed learn-to-grow approach, we show that tuning the ``{\tt reuse}" layers by using a simple $l_2$ based regularization on previous task parameters \CHANGE{(i.e. $\|\Theta_{i} - \Theta_{j}\|_2^2$, where $\Theta_i$ is the parameters for the current task and $\Theta_j$ is the  parameters from the $j$-th task that selected to reuse)} is sufficiently safe in terms of eliminating the forgetting. Both strategies, fixing the ``{\tt reuse"} layers or fine-tuning them with simple $l_2$ regularization  can keep the overall accuracy as high as training each task individually (see Fig. \ref{fig:mnist_result} (a)). \CHANGE{Encouraged by the above result, we further conduct experiments by tuning the ``{\tt reuse}" layers with smaller learning rate \emph{without using any regularization}. In other words, we do not add any regularization to the parameters to mitigate forgetting among different tasks. The results are shown in Fig.~\ref{fig:mnist_result} (a), which almost have the same behavior compared to the $\l_2$ regularization. This suggests that the learned structures actually make sense in terms of the ``{\tt reuse"} decisions, and the reused parameters are near-optimal for specific tasks.}


%
\begin{figure}[tbp!]
\centering
\includegraphics[width = 0.40\textwidth,trim={0 0.1in 0 0},clip]{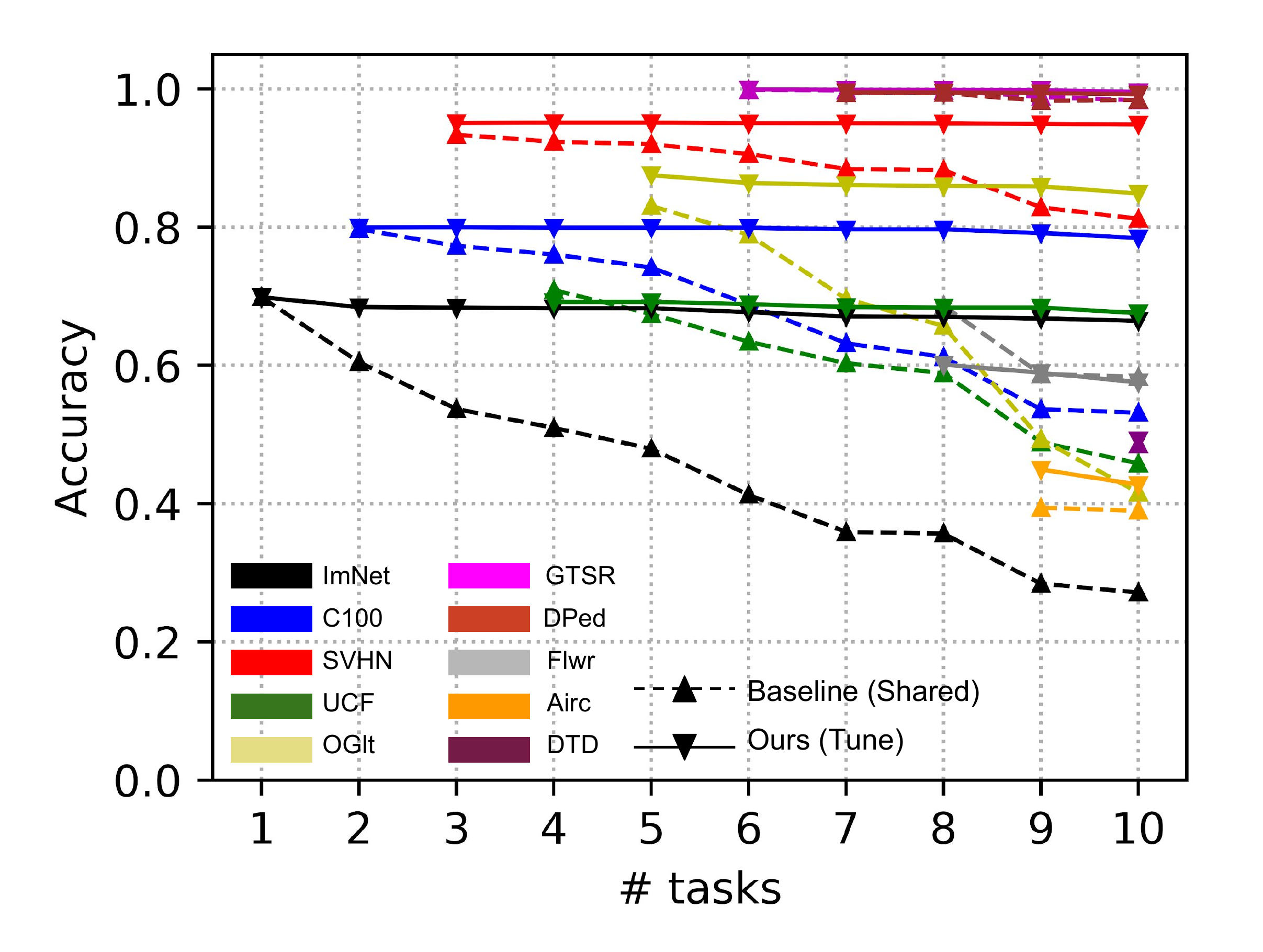}
\vspace{-10pt}
\caption{Comparisons of the catastrophic forgetting effects between our proposed approach and the baseline in VDD. }
\label{fig:vdd_result}
\end{figure}

We continue to test this in VDD. A predefined order of the ten tasks is used: ImageNet, CIFAR-100, SVHN, UCF101, Omniglot, GTSR, DPed, Flower, Aircraft and Textures (results for different ordering are provided in the supplementary material). As a baseline, we also train a model that shares all layers in the backbone and updates them from task to task. \CHANGE{The baseline model and our learn-to-grow  model are trained with similar settings as in the permuted MNIST experiments, and we choose not to use any regularization for fine-tuning the ``{\tt reuse"} layers due to the positive results we obtain in the permuted MNIST experiment.} As can be seen from Fig.\ref{fig:vdd_result}, our learn-to-grow method significantly outperforms the baseline approach. 

We also compare with other baselines in VDD and the results are shown in  Table.~\ref{table:vdd}. Our method obtains the best overall results and the total model size is similar to the ``adapter" approach~\citep{rebuffi2017learning}\footnote{The adapter proposed by Rebuffi et al. is targeted for the VDD dataset, which is not a continual learning method.}. Our approach obtains the best results in five out of nine tasks. Especially in tasks with small data size, e.g. VGG-Flowers and Aircraft, our method outperforms other baselines by a large margin. 
\begin{table*}[!tbph]
        \centering 
        \small{
        \resizebox{0.9\textwidth}{!}
        {{\renewcommand{\arraystretch}{1.3}%
        \begin{tabular}{l|cccccccccc|c|c}
        Model & ImNet & C100 & SVHN & UCF & OGlt & GTSR & DPed & Flwr & Airc. & DTD & avg. & \#params \\ \hline
        Individual & 69.84 & 73.96 & \textbf{95.22} & 69.94 & \textbf{86.05} & \textbf{99.97} & \textbf{99.86} & 41.86 & \textbf{50.41} & 29.88 & 71.70 & 58.96M \\ \hline 
        Classifier & 69.84 & 77.07 & 93.12 & 62.37 & 79.93 & 99.68 & 98.92 & \textbf{65.88} & 36.41 & 48.20 & 73.14 & 6.68M  \\ \hline
        Adapter & 69.84 & \textbf{79.82} & 94.21 & \textbf{70.72} & 85.10 & \textbf{99.89} & 
        \textbf{99.58} & 60.29 & 50.11 & \textbf{50.60} & \textbf{76.02} & 12.50M \\ \hline 
        Ours (fix) & 69.84 & \textbf{79.59} & \textbf{95.28} & \textbf{72.03} & \textbf{86.60} & 99.72 & 99.52 & \textbf{71.27} & \textbf{53.01} & \textbf{49.89} & \textbf{77.68} & 14.46M \\ \hline 
        \end{tabular} }}}
        \\ [1ex]
        \caption{Results of the (top-1) validation classification accuracy (\%) on Visual Domain Decathlon dataset, top-2 performance are highlighted. The total model size (``\#params") is the total parameter size (in Million) after training the 10 tasks. Individual indicates separate models trained for different tasks. Classifier denotes that a task specific classifier (i.e. the last softmax layer) is tuned for each task. \textbf{Adapter refers to methods proposed by Rebuffi et al.\cite{rebuffi2017learning}}. }\label{table:vdd} \vspace{-10pt}
\end{table*}

\begin{figure}[tbp!]
\centering
\includegraphics[width = 0.45\textwidth,trim={0 0.1in 0 0},clip]{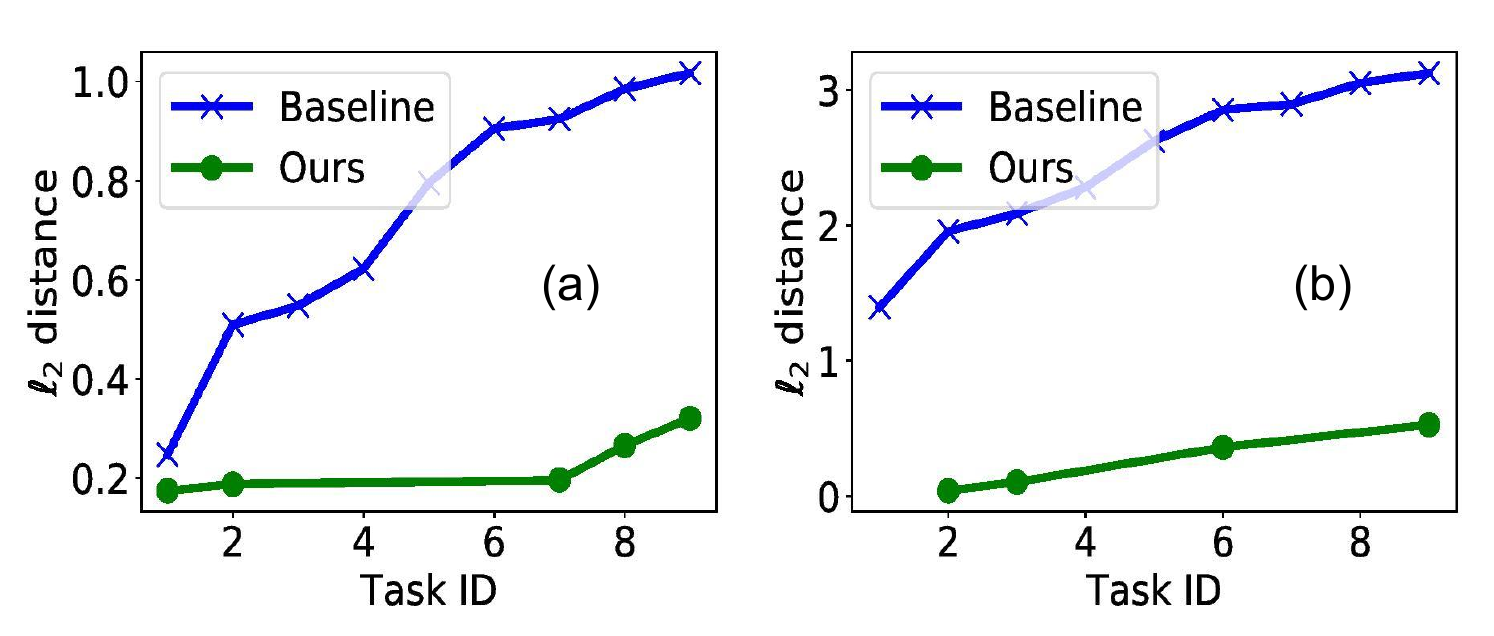}
\vspace{-10pt}
\caption{Distance between the tuned parameters at each task and the parameters of the very first task in VDD experiments. a) First layer parameter distance, and b) Last layer parameter distance. Baseline indicates the result from tuning all layers using SGD. }
\label{fig:distance}
\end{figure}

\CHANGE{To analyze why the simple fine-tuning strategies for the ``{\tt reuse}" layers work, we calculate the $l_2$ distance between the parameters before and after fine-tuning for each task in VDD. We want to check if the ``{\tt reuse}" layers are almost at an optimal position for the current task to use (i.e., the $l_2$ distance will be relatively small). Fig.~\ref{fig:distance} (a) and (b) show the $\ell_2$ distances between the parameters learned in the very first task and those after tuned in the following tasks for the first and last layers respectively. It is clear that the fine-tuned parameters in our learn-to-grow method do not move far away from the original location in the parameter space as compared to the baseline method, which explains why our method has less forgetting in fine-tuning the ``{\tt reuse"} layer\footnote{Similar trend of the distances between parameters across tasks was found for all layers. In general, the higher a layer is in the network the more the parameter moves for the baseline method, whereas for our learn-to-grow method the distances are typically very small.}. In addition, we notice that the distances in our methods are more or less at the same scale across all layers. This may attribute to the fact that the learn-to-grow of parameters and structures is explicitly optimized, and thus the selected ones are more compatible with a new  task. Therefore, less tuning is required for the new task and hence smaller distances.}

\CHANGE{Experimental results in this section show that the explicitly continual structure learning  is important. With the proper structures learned, all the relevant parameters from previous tasks can be exploited. Additionally, since the way to leverage these parameters are learned through the structure optimization, much less tuning is required for better performance on new tasks, and forgetting can thus be overcomed.}

\subsection{Comparisons with State-of-the-Art Methods}

\begin{figure}[!htbp]
\centering
\includegraphics[width = 0.50\textwidth,trim={0 0.1in -0.2in 0},clip]{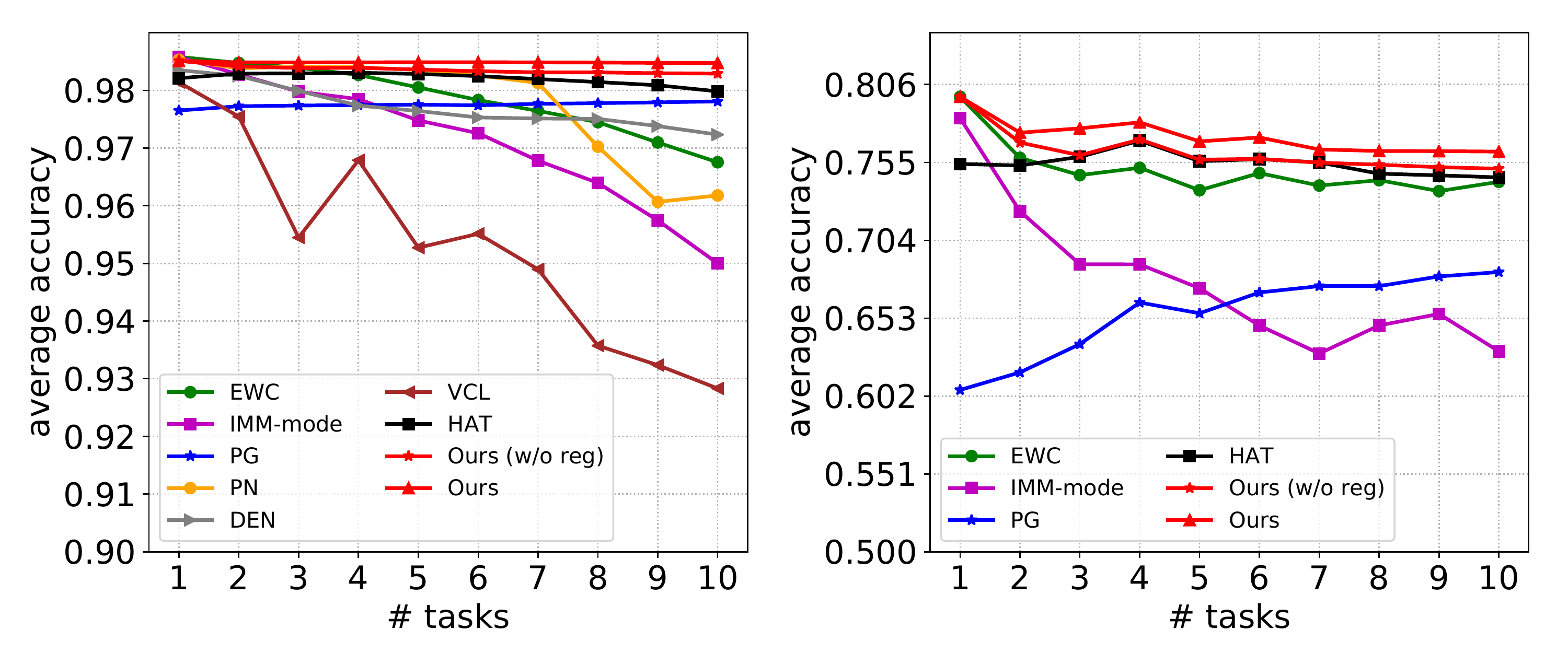}
\vspace{-15pt}
\caption{Performance comparisons in a) permuted MNIST and b) split CIFAR-100 dataset. Methods include 
    \citet[EWC]{kirkpatrick2017overcoming}, 
     \citet[IMM]{lee2017overcoming}, 
     \citet[PathNet (PN)]{fernando2017pathnet}, 
     \citet[Progressive Net (PG)]{rusu2016progressive}, 
     \citet[HAT]{serra2018overcoming},
     \citet[DEN]{lee2017overcoming},
     \citet[VCL]{nguyen2018variational}, ours (w/o reg) denotes the case where finetuning for current tasks is done without using any regularization to prevent forgetting, and ous represents the case where the $\ell_2$ regularization is used.}
\label{fig:pmnist_comparison}
\end{figure}

\CHANGE{We compare our learn-to-grow method with other recent continual learning approaches -- \citet[DEN]{lee2017overcoming}, \citet[HAT]{serra2018overcoming}, \citet[EWC]{kirkpatrick2017overcoming}, \citet[IMM]{lee2017overcoming},  \citet[ProgressiveNet]{rusu2016progressive}, \citet[PathNet]{fernando2017pathnet},
\citet[VCL]{nguyen2018variational}. We compare the performance of various methods in the permuted MNIST dataset with $10$ different permutations. Since our model adds more parameters, for fair comparisons we also train other methods with comparable or more parameters. In particular, since our model tends to add new parameters at the first layer, for all methods we increase the number of neurons in the first hidden layer by ten times, so that theoretically they could learn exactly the same structure as our model. 
We also tried to compare with \citet{shin2017continual}, however, we are unable to get reasonable performance, and hence the results are not included. 
The results are shown in Fig. \ref{fig:pmnist_comparison} (a) and Table~\ref{table:pmnist}. It is clear that our method, either tuned with or without regularization, performs competitive or better than other methods on this task. This result suggests that although theoretically, structure can be learned along with parameter, in practice, the SGD-based optimization schema have a hard time achieving this. This in turn indicates the importance of explicitly taking continual structure learning into account when learning tasks continuously.}
\CHANGE{Although both DEN and our method dynamically expand the network, our performance is much better, which is attributed to the ability of learning new structures for different tasks. Additionally, our model performs competitive as or better than the methods that completely avoids forgetting by fixing the learned weights, such as ProgressNet and PathNet,  without enforcing such restrictions.}

\CHANGE{We further compare with other methods in the split CIFAR-100 dataset~\citep{lopez2017gradient}, where we randomly partition the classes of CIFAR-100 into $10$ disjoint sets, and regard learning each of the 10-class classification as one task. Different from the permuted MNIST, the split CIFAR-100 presents a continual learning scenario  where the input distribution is similar (i.e., natural images) whereas the output distribution is different (disjoint classes).  We choose to use Alexnet~\citep{AlexNet} as the network structure, and all methods are constrained to have comparable model complexities. This network structure contains three convolution and max pooling layers and two fully connected layers before the last classification layer. Comparison results are shown in Fig~\ref{fig:pmnist_comparison} (b) and Table~\ref{table:cifar100}. Similar results as the MNIST experiment are obtained in this experiment. Interestingly, for all tasks, our method always seeks the structures that use new parameters for the last convolution layer and reuse the rest of the network parameters. It makes sense since the lower layer features are shared accounting for the similar input distribution, and the higher ones need to be specific for different tasks due to different output distribution. The fully connected layers are all selected to be ``{\tt reused}" instead of ``{\tt new}", and this may be because of the relatively large capacity that is sufficiently powerful to learn the subsequent tasks. }

\begin{table}[ht]
    \centering
    \small{
    \resizebox{0.35\textwidth}{!}{
    \begin{tabular}{c|c|c}
    \hline
    Method & Acc (\%) & \#params \\ \hline
    SGD & 72.51 & 3.35M \\ 
    EWC & 96.75 & 3.35M  \\
    IMM & 95.00 & 3.35M  \\
    VCL & 95.32 & 3.35M \\
    HAT & 97.98 & 3.46M \\ 
    PN & 96.18 & 3.96M \\
    PG & 97.81 & 3.05M \\
    DEN & 97.71 & 3.53M \\\hline
    Ours (fix) & 98.46 & 2.56M \\
    Ours (tune) & 98.29 & 2.56M \\
    Ours (tune+L2Reg) & 98.48 & 2.56M \\\hline
    \end{tabular} }}
    \\ [1ex]
    \caption{Results of different continual learning approaches on 10 permutated MNIST datasets. The averaged accuracy after all 10 tasks are learned and total number of parameters are compared. 
    }\label{table:pmnist} 
\end{table}

\begin{table}[ht]
    \centering
    \small{
    \resizebox{0.35\textwidth}{!}{
    \begin{tabular}{c|c|c}
    \hline
    Method & Acc (\%) & \#params \\ \hline
    SGD & 21.02 & 6.55M\\ 
    EWC & 74.23 & 6.55M \\
    IMM & 63.13 & 6.55M \\
    HAT & 74.52 & 6.82M \\ 
    PN & 60.48 & 7.04M \\
    PG & 68.32 & 6.80M \\\hline
    Ours (fix) & 75.31 & 6.86M \\
    Ours (tune) & 75.05 & 6.86M \\
    Ours (tune+L2Reg) & 76.21 & 6.86M \\\hline
    \end{tabular} }}
    \\ [1ex]
    \caption{Results of different continual learning approaches on split CIFAR100 dataset. The averaged accuracy after all 10 tasks are learned and total number of parameters are compared.
    }\label{table:cifar100} 
\end{table}


\section{Related Work}
\label{others}
Continual learning~\citep{thrun1995lifelong} remains a long standing  problem, as models have the tendency to forget previously learned knowledge when trained on new information~\citep{thrun1995lifelong, mccloskey1989catastrophic}. This is referred as the catastrophic forgetting problem in the literature. Early attempts to alleviate catastrophic forgetting include memory systems that store previous data and replay the stored old examples with the new data~\citep{robins1995catastrophic}, and similar approaches are still used in the latest development~\citep{rebuffi2017icarl, li2018supportnet, lopez2017gradient}. \CHANGE{\citet{shin2017continual} proposes to learn a generative model to capture the data distribution of previous tasks, and both generated samples and real samples from the current task are used to train the new model so that the forgetting can be alleviated.}

On the one hand, a typical class of methods for mitigating catastrophic forgetting relies on regularization which imposes constraints on the update of model parameters. 
\citet{kirkpatrick2017overcoming} proposes elastic weight consolidation (EWC) whose objective is to minimize the change of weights that are important to previous tasks through the use of a quadratic constraint.  \citet{zenke2017continual} proposes to alleviate catastrophic forgetting by allowing individual synapse to estimate their importance for solving learned tasks, then penalizing the change on the important weights. \CHANGE{\citet{schwarz2018progress} presents a method that divides the learning into two phases -- progress and compress. During the progress phase, the model makes use of the previous model for learning the new task. In the compress phase, the newly learned model is distilled into the old model using EWC to alleviate forgetting. \citet{serra2018overcoming} proposes to use attention mechanism to preserve performance for previous tasks.} Other methods could also completely avoid forgetting by preventing changes to previous task weights~\citep[see for example][]{rusu2016progressive,mallya2018piggyback,fernando2017pathnet}.

On the other hand, another class of methods for continual learning is allowing the model to expand. Dynamically expandable networks~\citep{lee2017lifelong} select whether to expand or duplicate layers based on certain criteria for a new task. However, the model for the new task is forced to use the old structure from previous tasks. \CHANGE{Similar strategies are adopted in progressive networks~\citep{rusu2016progressive}. Our proposed learn-to-grow framework is more flexible, thanks to the structure optimization via NAS. PathNet~\citep{fernando2017pathnet} selects paths between predefined modules, and tuning is allowed only when an unused module is selected. Our method  does not have any restriction on tuning parameters from previous tasks.}

Our method also relates to neural architecture search~\citep{stanley2002evolving,zoph2016neural,baker2017accelerating,liu2018darts}, as we employ search methods to implement the structure optimization. In particular, DARTS~\citep{liu2018darts} is used for efficiency, where a continuous relaxation for architecture search is proposed. 

\section{Conclusion}
In this paper, we present a simple yet effective learn-to-grow framework for overcoming catastrophic forgetting in continual learning with DNNs, which explicitly takes into account continual structure optimization via differentiable neural architecture search. We introduce three intuitive options for each layer in a give model, that is to ``{\tt reuse}", ``{\tt adapt}" or ``{\tt new}" it in learning a new task.  In experiments, we observed that the explicit learning of model structures leads to sensible structures for new tasks in continual learning with DNNs. And, catastrophic forgetting can be either completely avoided if no fine-tuning is taken for the ``{\tt reuse}" layers, or significantly alleviated even with fine-tuning. The proposed method is thoroughly tested in a series of datasets including the permuted MNIST dataset, the visual decathlon dataset and the split CIFAR-100 dataset. It obtains highly comparable or better performance in comparison with state-of-the-art methods.   

\section*{Acknowledgements}
X. Li and T. Wu are supported by ARO grant W911NF1810295 and DoD DURIP grant W911NF1810209. T. Wu is also supported by NSF IIS 1822477. The authors are grateful for all the reviewers for their helpful comments.

\nocite{langley00}

\bibliography{example_paper}
\bibliographystyle{icml2019}

\end{document}


\twocolumn[
\icmltitle{Supplementary Material -- Learn to Grow: A Continual Structure Learning Framework for Overcoming Catastrophic Forgetting}



\icmlsetsymbol{equal}{*}
\icmlsetsymbol{intern}{$\dagger$}

\begin{icmlauthorlist}
\icmlauthor{Xilai Li}{equal,intern,to}
\icmlauthor{Yingbo Zhou}{equal,sf}
\icmlauthor{Tianfu Wu}{to}
\icmlauthor{Richard Socher}{sf}
\icmlauthor{Caiming Xiong}{sf}
\end{icmlauthorlist}

\icmlaffiliation{to}{Department of Electrical and Computer Engineering and the Visual Narrative Initiative, North Carolina State University, NC, USA.}
\icmlaffiliation{sf}{Salesforce Research, Palo Alto, CA, USA.}

\icmlcorrespondingauthor{C. Xiong}{cxiong@salesforce.com}
\icmlcorrespondingauthor{T. Wu}{tianfu\_wu@ncsu.edu}

\icmlkeywords{Learn-to-Grow, Continual Learning, Lifelong Learning, Catastrophic Forgetting, Deep Neural Networks, Machine Learning, ICML}

\vskip 0.3in
]



\printAffiliationsAndNotice{\icmlEqualContribution} 

\setlength{\textfloatsep}{6pt}

\section{Additional Experimental Details for permuted MNIST}
For all MNIST experiment, we use fully connected layer with three hidden layer, each with 300 hidden units, and one shared output layer for our method. For all other methods except pathnet and progressive net we used 3000 units in the first layer and 300 for the rest. For pathnet, each module in the first layer has 300 units, and the result layers has 30 units. We use 16 modules per layer, and 5 layers for pathnet, and restrict each mutation to only use 3 modules besides the output layer. For progressive net, the first layer has 300 units for each task, and the rest layers each has 30 units. Therefore, all competitive methods are having more or the same number of parameters as our methods.

For variational continual learning~\citep[VCL][]{nguyen2018variational}, we used the official implementation at \url{https://github.com/nvcuong/variational-continual-learning}. For fair comparison with other methods, we set the coreset size to zero for VCL.

For \citep{shin2017continual} we used implementation from \url{https://github.com/kuc2477/pytorch-deep-generative-replay}. We tried various hyper-parameter settings, however, we are unable to get reasonable results on permutated MNIST. Performance was reasonable when the number of tasks is within five (average performance at around 96\%). When number of tasks go beyond five, performance drops on previous tasks is quite significant. Reaching 60\%

For DEN we use the official implementation at \url{https://github.com/jaehong-yoon93/DEN}, and we used \citet{serra2018overcoming} implementation of HAT, EWC, and IMM at \url{https://github.com/joansj/hat}. We used our own implemention for for Progressive Network and PathNet. All methods are trained using the same permutations and same subset of training data.

\section{Additional Experimental Details for Split CIFAR-100}
For all CIFAR-100 experiment, we use an Alexnet like structure. It contains three convolution and max pooling layers followed by two fully connected layers. The convolution layers are of size (4,4), (3,3) and (2,2) with 64, 128 and 256 filters, respectively. All convolution layers are followed by max pooling layer of size (2,2) and rectified linear activations. The two fully connected layers each have 2048 hidden units.

\section{Additional Experiments on Visual Decathlon Dataset}
\begin{table*}[ht]
        \centering 
        \small{
        \resizebox{\textwidth}{!}
        {{\renewcommand{\arraystretch}{1.3}%
        \begin{tabular}{l|c|cccccccccc|c}
         & & ImNet & C100 & SVHN & UCF & OGlt & GTSR & DPed & Flwr & Airc. & DTD & Tot.\\ \hline
        \multirow{2}{*}{$\beta=0.01$} & acc & 69.84 & 78.50 & 95.33 & 72.50 & 86.41 & 99.97 & 99.76 & 66.01 & 51.37 & 50.05 & 76.97\\ 
        & \#params & 6.07 & 0.15 & 2.74 & 2.28 & 6.17 & 3.59 & 1.02 & 0.19 & 4.15 & 0.13 & 26.49 \\ \hline 
        \multirow{2}{*}{$\beta=0.1$} & acc & 69.84 & 79.59 & 95.28 & 72.03 & 86.60 & 99.72 & 99.52 & 71.27 & 53.01 & 49.89 & 77.68 \\ 
        & \#params & 6.07 & 0.34 & 1.19 & 1.32 & 3.19 & 0.02 & 0.27 & 0.16 & 1.86 & 0.04 & 14.46\\ \hline
        \multirow{2}{*}{$\beta=1.0$} & acc & 69.84 & 78.00 & 93.40 & 63.83 & 84.30 & 99.78 & 99.01 & 65.77 & 39.27 & 48.77 & 74.20 \\ 
        & \# params & 6.07 & 0.04 & 0.03 & 0.12 & 0.66 & 0.02 & 0.01 & 0.02 & 0.35 & 0.02 & 7.34 \\ \hline 
        \end{tabular} }}}
        \\ [1ex]
        \caption{Comparison of (top-1) validation classification accuracy (\%) and total model size (in Million) on Visual Domain Decathlon dataset with parameter loss factor $\beta$ of 0.01, 0.1, 1.0. }\label{table:vdd_model_size} 
\end{table*}
In the multi-task continual learning experiments, the 10 tasks was trained in a random sequence except the first 
task was fixed to be ImageNet. This is just for
fair comparison with other works such as \citet{rebuffi2017learning} and \citet{mallya2018piggyback}, they are all using a light weight module to adapt ImageNet
pretrained model to other of the 9 tasks. In real case, the tasks can come in any order, thus our framework would
be much more flexible. As the tasks are trained in sequence, a super model is maintained that all the newly
created weights and task-specific layers are stored. In this ResNet-26 model, all the Batch Normalization (BN) layers 
are treated as task-specific, which means each task has its own sets of BNs. Here, we fixed the weight during retraining 
when "reuse" is selected in the search phase. 
This means that the results of previous tasks would not be affected, i.e. no forgetting. We leave the evaluation
of forgetting in the context of VDD dataset as future work. 

In Table 1, we compare the results using our approach with other baselines. "Individual" means that each task is trained
individually and weights are initialized randomly. "Classifier" means that only the last layer classifier could be tuned while the former 25 layers 
are transfer from ImageNet pretrained model and kept fixed during training. In this case, each task only adds a task-specific
classifier and BNs, thus the overall model size is small. "Adapter" add a 1x1 conv layer besides each 3x3 conv layer, 
and the outputs will be added before proceed to the next layer. Due to the lightweight 1x1 conv layer, each task will
add approximately 1/9 of the whole model size. As shown in table 1, the results achieved by our framework is better than
other baselines and the total model size is similar to "Adapter" case. We can see that our approach gives best results in five out of nine tasks. Especially in task with small data size, e.g. VGG-Flowers and Aircraft, our method outperforms
other baselines by a large margin. 

Due to each choice has different parameter cost, we add a parameter loss function to $L_{val}$ to penalize the choices 
that cost additional parameters.  And the value of the loss function is proportional to the product of the additional 
parameter size and its corresponding weight value $\alpha_c^l$. In table 2, we test it with three different scaling
factor $\beta$ of the parameter loss. We found that the scaling factor $\beta$ can control the additional parameter
size for each task. And we find that $\beta=0.1$ gives the best average accuracy and can control the total model size
approximate $2.3\times$ compared with the original model size. 

\begin{figure}[tbp!]
\centering
\includegraphics[width = 0.45\textwidth,trim={0 0.1in 0 0},clip]{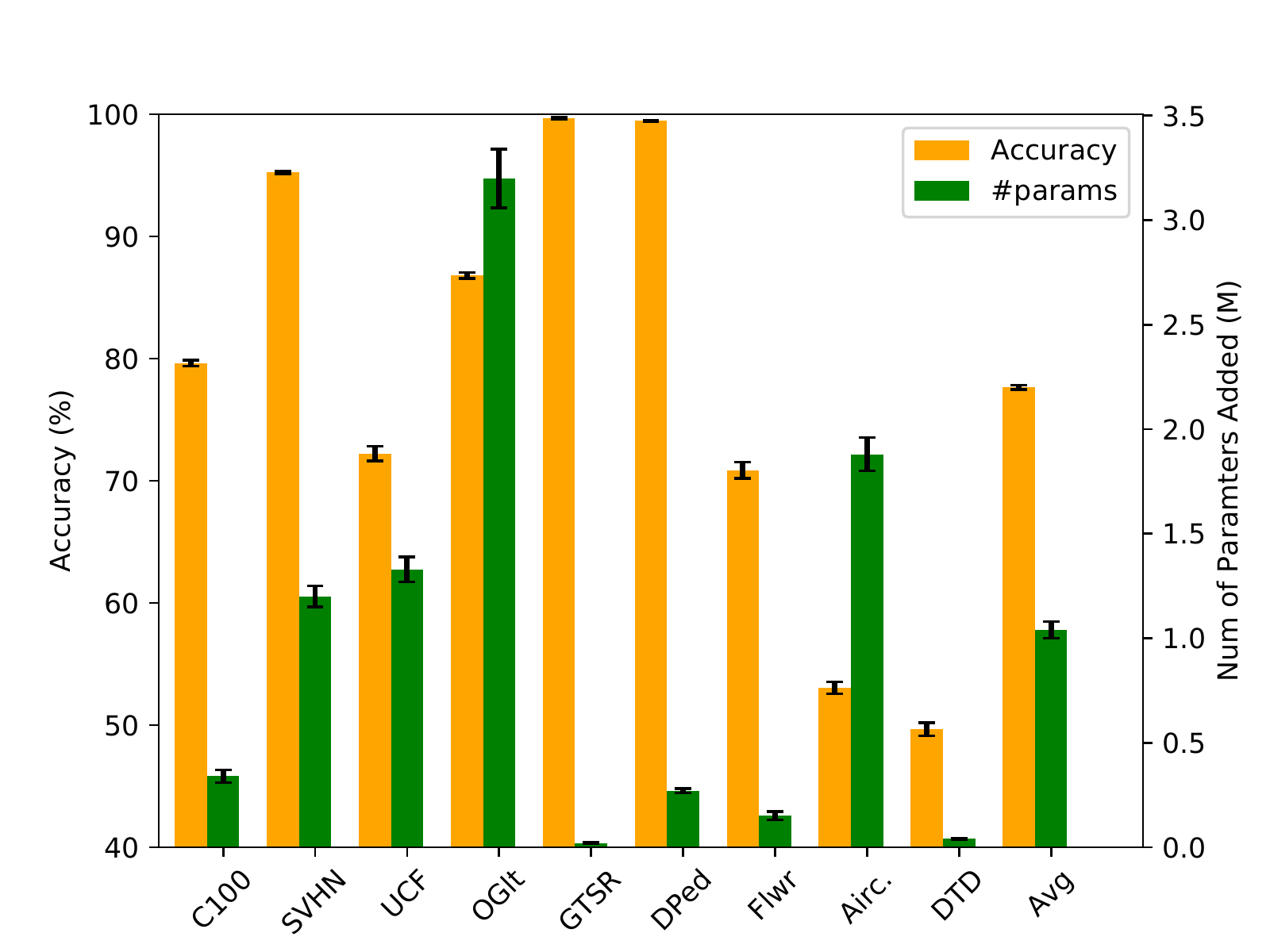}
\vspace{-10pt}
\caption{Statistics for performance and number of added parameters for each task of VDD dataset with 4 random task ordering. The first task is kept with ImageNet due to its large size and long training time. We observed that both accuracy and parameter growth are robust to different task ordering.}
\label{fig:distance}
\end{figure}


\nocite{langley00}

\bibliography{example_paper}
\bibliographystyle{icml2019}